\newif\ifreview
\documentclass{article}
\usepackage{ijcai26}
\usepackage{times} 
\usepackage{soul}
\usepackage{url}
\usepackage[hidelinks]{hyperref}
\usepackage[utf8]{inputenc}
\usepackage[small]{caption}
\usepackage{graphicx}
\usepackage{subfigure}
\usepackage{subcaption}
\usepackage{placeins}
\usepackage{amsmath}
\usepackage{amsthm}
\usepackage{booktabs}
\usepackage{algorithm}
\usepackage{algorithmic}
\usepackage{xcolor}
\usepackage[switch]{lineno}
\usepackage{placeins}
\newif\ifanonymize
\newcommand{\anon}[2]{\ifanonymize #1\else #2\fi} 
\ifreview
    \linenumbers
\fi
\definecolor{darkgreen}{rgb}{0, 0.6, 0} 

\title{A Resilient Solution for Sewer Overflow Monitoring Across Cloud and Edge}
\anon{\author{\textbf{Anonymous Author(s)}}}
{
    \author{
    Vipin Singh$^1$
    \and
    Tianheng Ling$^2$\and
    Peter Ghaly$^3$\and \\
    Felix Grimmeisen$^3$\and
    Gregor Schiele$^2$ \And
    Felix Biessmann$^{1,4}$\\
    \affiliations
    $^1$Berlin University of Applied Sciences, Berlin, Germany\\
    $^2$University of Duisburg-Essen, Duisburg, Germany\\
    $^3$Okeanos Smart Data Solutions GmbH, Bochum, Germany\\
    $^4$Einstein Center Digital Future, Berlin, Germany\\
    \emails
    \{vipin.singh, felix.biessmann\}@bht-berlin.de,
    \{tianheng.ling, gregor.schiele\}@uni-due.de, \\
    \{peter.ghaly, felix.grimmeisen\}@okeanos.ai
    }
}
\begin{document}
\maketitle


\begin{abstract}
Aging combined sewer systems in many historical cities are increasingly stressed by extreme rainfall events, which can trigger combined sewer overflows (CSO) with significant environmental and public health impacts. Forecasting the filling dynamics of overflow basins is critical for anticipating capacity exceedance and enabling timely preventive actions for CSO. We present a web-based demonstrator that integrates Deep Learning forecasting methods in both cloud and edge settings into an interactive monitoring dashboard for overflow monitoring, resilient to network outages.
\end{abstract}

\section{Introduction}

Combined Sewer Systems (CSS), which transport stormwater and wastewater through shared pipe networks, remain common in many historical cities but are increasingly stressed by extreme rainfall events driven by climate change~\cite{allard2021climate,ritchie2023urbanization,wilbanks2014climate}. During heavy precipitation, CSS can exceed their hydraulic capacity, leading to Combined Sewer Overflows (CSO) that discharge untreated wastewater into the environment with serious ecological and public health consequences~\cite{bolan2023impacts,botturi2021combined}. To mitigate peak loads, many cities adopt overflow basins as temporary buffers~\cite{baneerjee2024overflow}. However, without reliable estimates of current and future filling-levels, it remains difficult to trigger timely preventive actions such as flow diversion or operational adjustments before critical thresholds are reached. Forecasting filling dynamics has therefore become a key requirement for proactive CSS management under increasingly volatile weather conditions.

Recent work has shown that data-driven forecasting models can support CSO management in both centralized cloud settings with full sensor access~\cite{singh_data-driven_nodate,singh_evaluating_2025} and decentralized edge deployments designed to tolerate network or infrastructure failures~\cite{ling_automated_2025}. These approaches, however, are typically developed and evaluated under separate deployment assumptions. In practice, sewer monitoring systems must operate across different conditions where connectivity, sensor availability, and computational resources can change over time.

Building on prior cloud-based and edge-based CSO forecasting studies, this work consolidates these capabilities into a practical, interactive monitoring framework for public-interest infrastructure management~\cite{zueger2023piai}. The proposed web-based demonstrator enables exploration of early-warning signals and overflow risk indications derived from filling-level forecasts under both standard cloud operation and constrained edge-based operation during network outages. Our demonstrator supports transparent risk communication for sewer system operators, municipal infrastructure managers, and other stakeholders involved in urban water management, facilitating informed and accountable decision-making in safety-critical settings.


\section{System Framework}

This system supports timely risk awareness and preventive decisions by integrating cloud- and edge-based forecasting into simulated operational settings. Two deployment assumptions reflect realistic monitoring conditions (see \autoref{fig:system_overview}):
\begin{figure}[!bp]
    \centering
    \includegraphics[width=\linewidth]{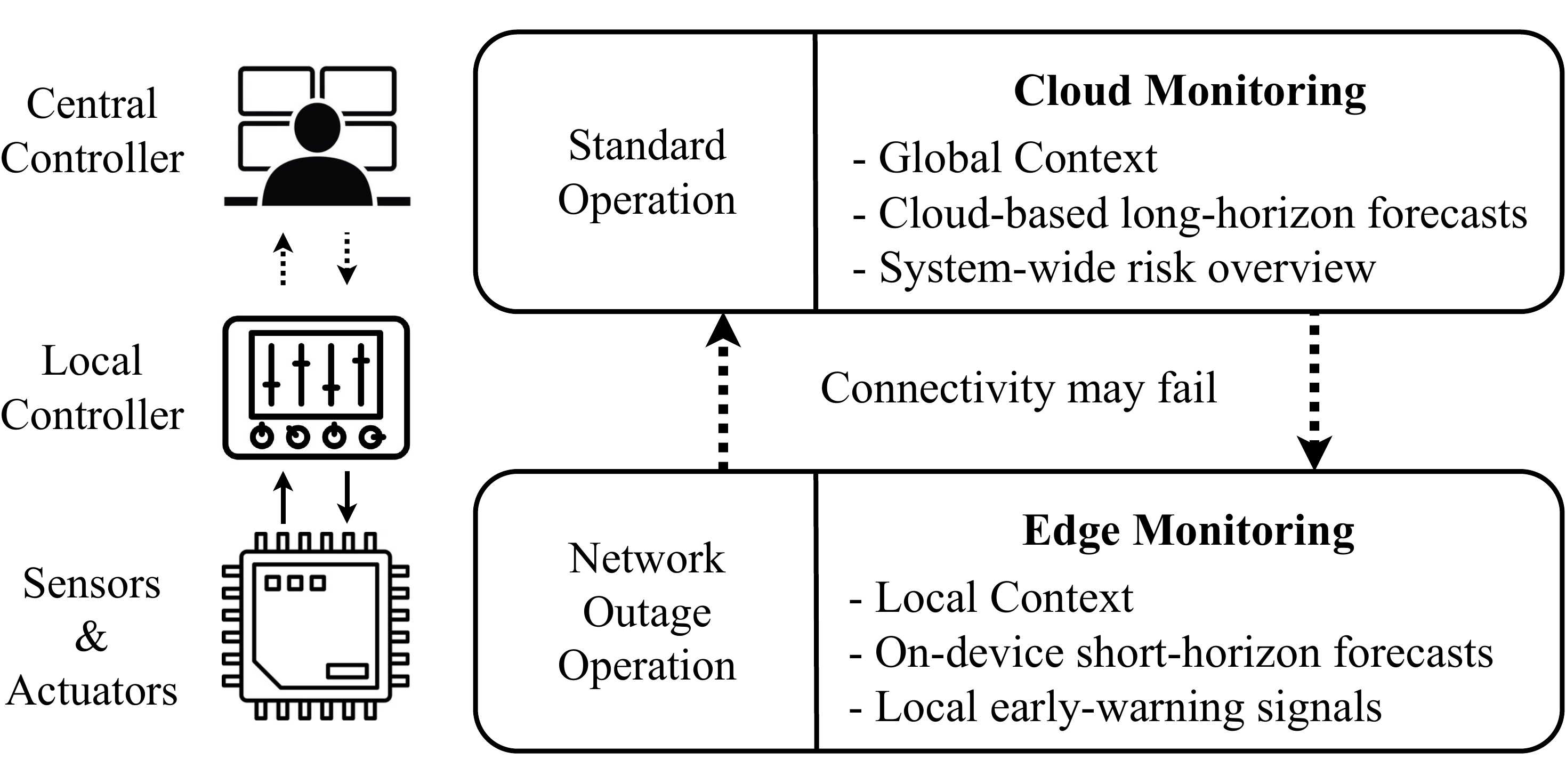}
    \caption{System architecture highlighting operational redundancy through cloud-based and edge-based monitoring and forecasting.}
    \label{fig:system_overview}
\end{figure}
\begin{itemize}
    \item \textbf{Under standard operation}, stable cloud connectivity and full sensor availability are assumed. In this mode, filling-level forecasts are computed on the cloud using a global view of the system state to facilitate both short- and long-term system planning. The cloud-based models integrate measurements from various sensors capturing diverse aspects of CSS operation. These models then generate long-horizon forecasts of future filling-levels over a 12-hour prediction window to support proactive planning. In addition, we incorporate an overflow risk estimation for 2-hour-ahead early warning.

   \item \textbf{Under network outage operation}, cloud communication is assumed to be (temporarily) unavailable. It emulates an edge-based environment in which filling-level forecasts are computed locally at the monitoring node. Forecasting relies exclusively on historical filling-level measurements from the target overflow basin and produces short-horizon forecasts with a one-hour prediction window to maintain early-warning functionality under constrained conditions.
\end{itemize}
This design maintains early-warning and system-planning across changing conditions while enabling systematic comparison of filling-level dynamics between cloud- and edge-based solution within a shared temporal and data context.

\section{Technical Implementation}

The demonstrator is displayed as a web-based dashboard organized into two types of modules: (1) control modules, which define the operational context and temporal progression of the simulation, and (2) monitoring modules, which expose sensor states, filling-level forecasts, rainfall measurements, and forecasts, and overflow risk assessments derived from the selected configuration. 
The demonstrator is realized as a \textit{Streamlit} application that loads and visualizes precomputed data locally. It is containerized and deployed on a compute cluster at the Berlin University of Applied Sciences, ensuring portability and reproducibility of the demo environment. The live demonstrator is publicly accessible\footnote{\UrlFont{https://riwwer.demo.calgo-lab.de}\label{demo-link}}.

\subsection{Data Source}

This study uses measurements collected from a large CSS in Duisburg, Germany. The dataset was provided by the municipal wastewater utility Wirtschaftsbetriebe Duisburg and covers six locations within the Vierlinden district, comprising a total of 35 sensors. These sensors capture diverse aspects of CSS operation, including overflow basin filling-level, rainfall measurements, pump energy consumption, valve states, and rainfall. All sensor signals are hourly resampled and span a three-year period. Data from 2021--2022 are used for model training, while measurements from 2023 are reserved for evaluation and visualization within the dashboard. 

The demonstrator analyzes and visualizes the filling level trajectory of one overflow basin, located at the sewage treatment facility, under the two operational scenarios. Depending on the scenario, the forecasting context differs. Under standard operation, cloud-based forecasting utilizes up to 72 hours of historical measurements together with global sensor information. Under network outage operation, edge-based forecasting relies exclusively on the most recent 24 hours of local filling-level measurements from the target basin.

\subsection{Control Modules}

\begin{figure*}[!tbp]
    \centering
    \includegraphics[width=0.9\textwidth]{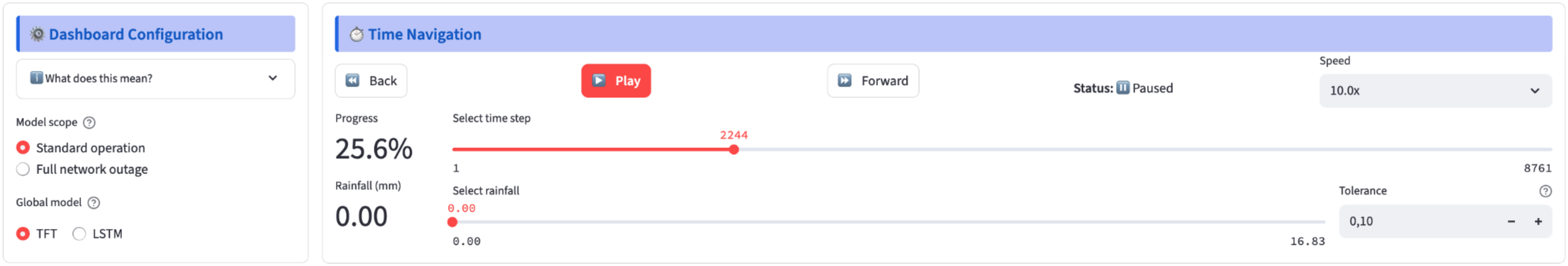}
    \caption{Control modules of the dashboard: dashboard configuration (left) and time navigation (right).}
    \label{fig:control_modules}
\end{figure*}

\autoref{fig:control_modules} illustrates the two control modules of the dashboard. The dashboard configuration module (see left panel in \autoref{fig:control_modules}) allows users to specify the operational assumptions under which early-warning signals are generated. Users can switch between \emph{standard operation} and \emph{full network outage}. In addition, this module enables the selection of the forecasting model architecture. The time navigation module (right in \autoref{fig:control_modules}) controls the temporal flow of the simulation. Users can replay historical data at adjustable speeds (at 0.1-15 hours per second), step forward or backward in time, or pause the simulation at specific timestamps. A dedicated rainfall-based navigation option allows direct access to time periods with specifiable precipitation intensities, facilitating targeted exploration of extreme rainfall events.

\subsection{Monitoring Modules}

The monitoring modules present sensor states, forecasts, and risk indicators derived from the selected operational configuration. These modules translate data availability and model outputs into interpretable visual cues, enabling users to understand early-warning signals and their underlying context under different deployment conditions.

\begin{figure}[!tbp]
\centering
\begin{minipage}[t]{0.48\columnwidth}
    \centering
    \includegraphics[height=0.18\textheight]{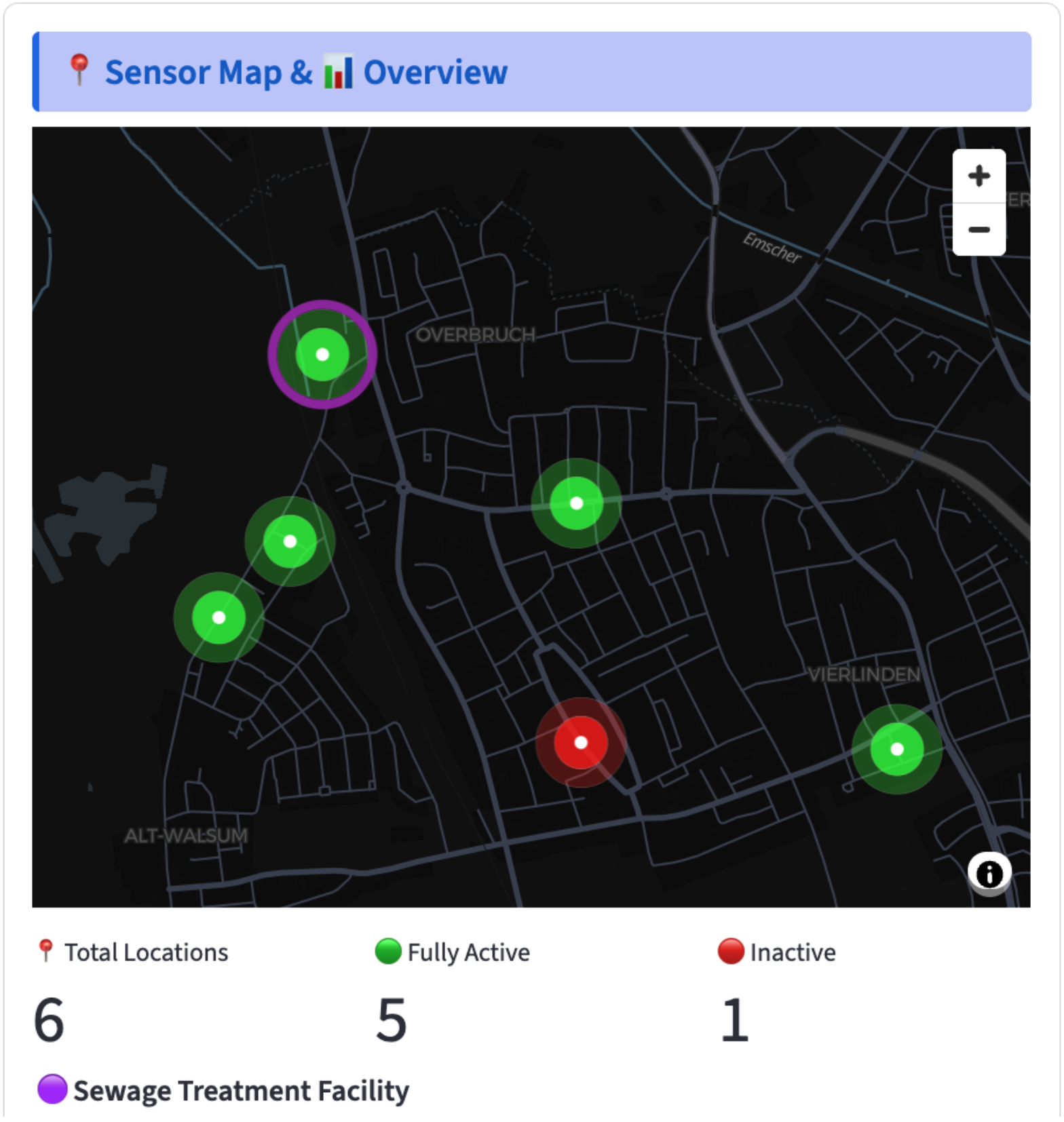}\\
    \small (a) Standard operation.
\end{minipage}
\hfill
\begin{minipage}[t]{0.48\columnwidth}
    \centering
    \includegraphics[height=0.18\textheight]{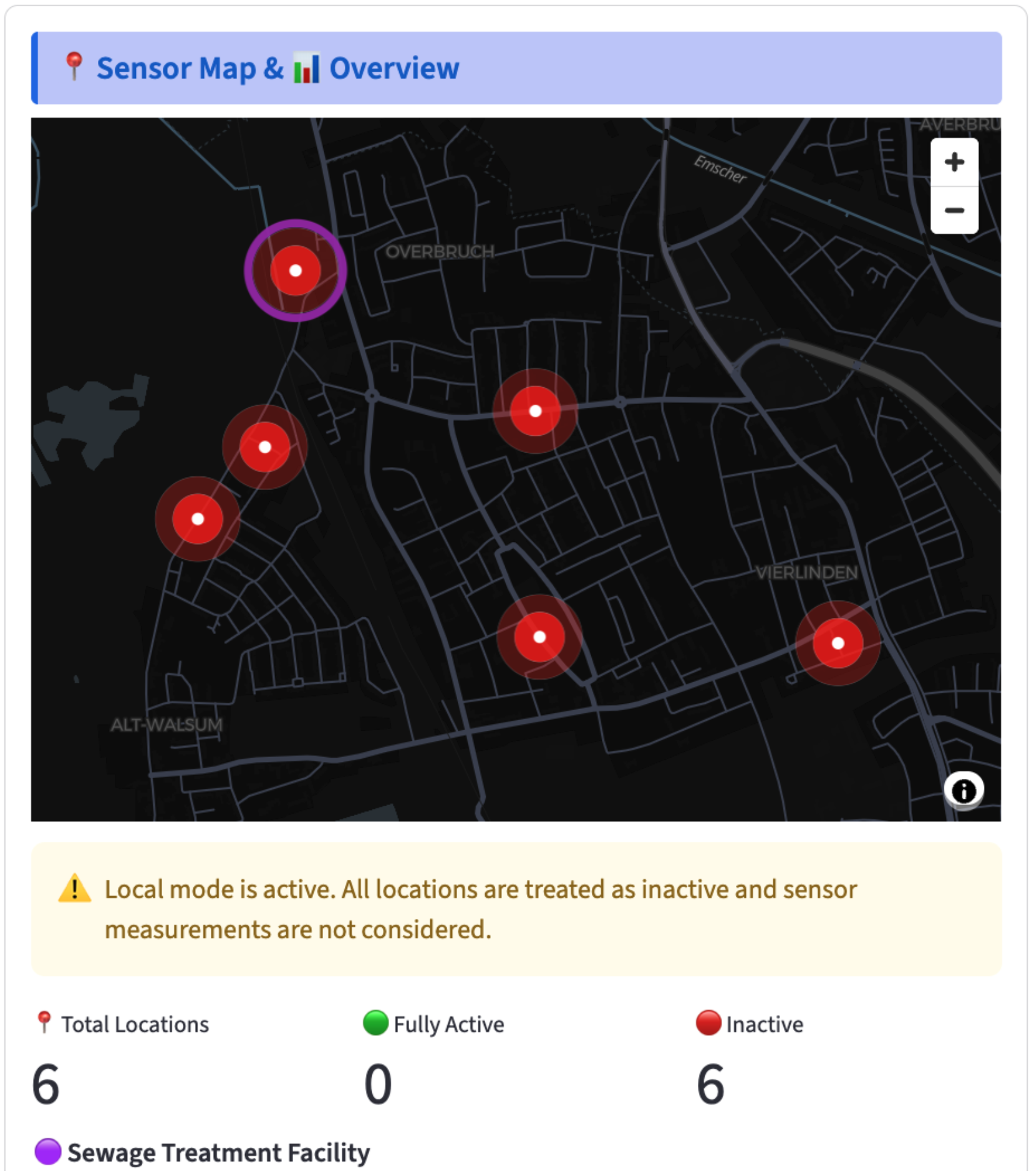}\\
    \small (b) Network outage.
\end{minipage}
\caption{Sensor availability by site: green markers show active sensors, red show inactive, and purple denotes the overflow basin.} 
\label{fig:sensor_status}
\end{figure}

\paragraph{Sensor Location and Status.}
As shown in \autoref{fig:sensor_status}, the sensor location and status module provides a spatial overview of the sewer network by visualizing sensor availability on an interactive geographic map. 
Each marker represents a monitoring location and indicates whether its associated sensors are active or inactive. Under standard operation, stable network connectivity and normal sensor operation are assumed. Most sensors are active, while missing values are filled using linear interpolation. To support long-horizon prediction and system-wide risk assessment, cloud-based forecasting aggregates available measurements into a global context. However, as displayed in \autoref{fig:sensor_status}(b), during network outage operation, connectivity to the cloud is assumed to be unavailable, and all remote locations are therefore treated as inactive. For instance, extreme weather events can increase the probability of such network disruptions.

\paragraph{Filling-Level Forecasting.}

\begin{figure}[!tbp]
\centering
\begin{minipage}[t]{0.48\columnwidth}
    \centering
    \includegraphics[width=\linewidth]{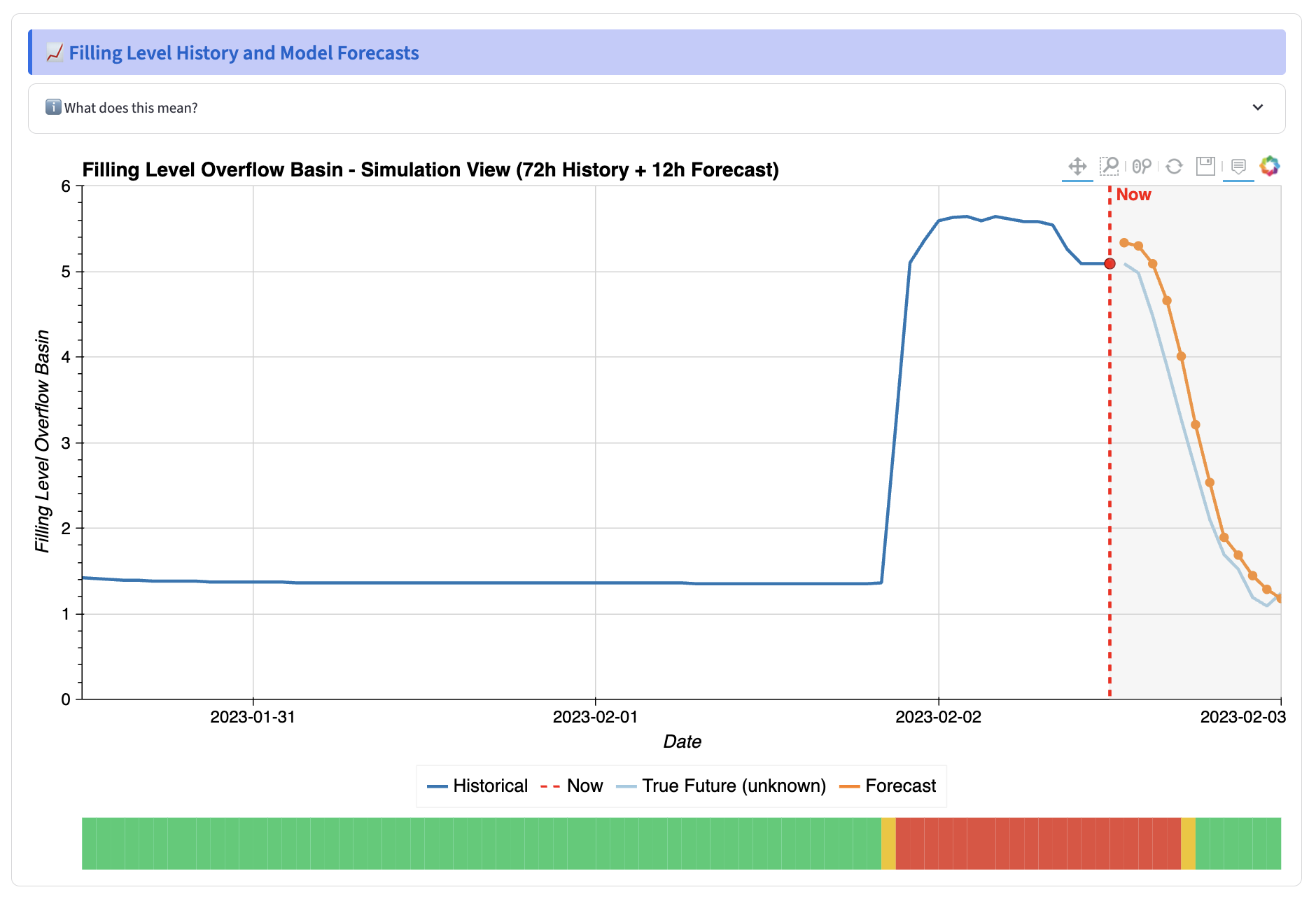}\\
    \small (a) Standard operation.
\end{minipage}
\hfill
\begin{minipage}[t]{0.48\columnwidth}
    \centering
    \includegraphics[width=\linewidth]{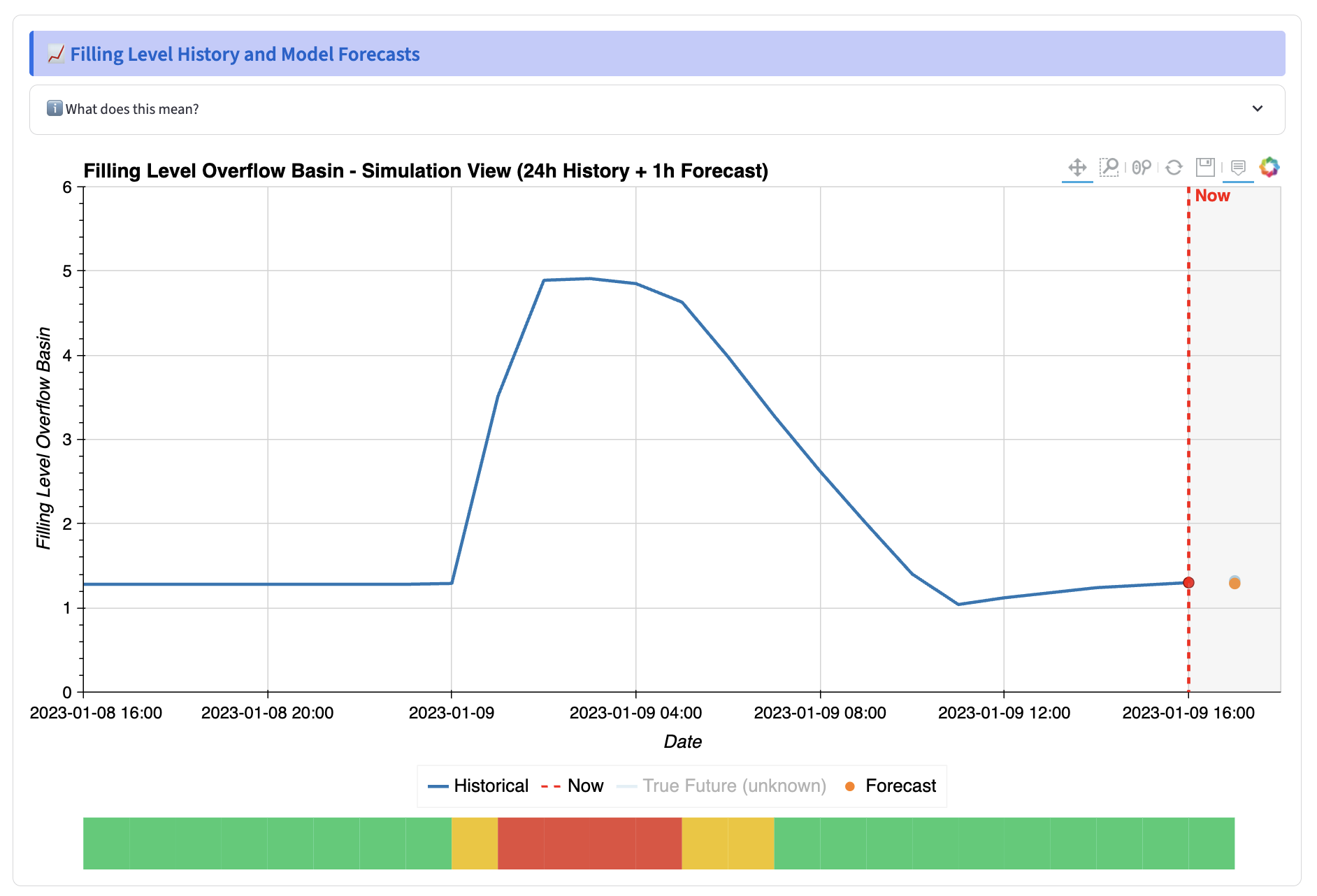}\\
    \small (b) Network outage.
\end{minipage}
\caption{Filling-level forecasts across operation conditions.}
\label{fig:filling_forecast}
\end{figure}

The filling-level forecasting module presents historical filling-level observations together with model predictions under the selected operational mode, as displayed in \autoref{fig:filling_forecast}. The visualization encodes observed values as a historical line, model predictions as an orange forecast line or marker depending on the condition, and the realized future trajectory as a light blue reference curve.
To complement the plot, an aligned, color-coded bar is positioned below the trajectory to categorize filling-level: low ($<$3 m, green), medium (3–4 m, yellow), and high ($>$4 m, red). Under standard operation, the dashboard shows long-horizon cloud-based forecasts generated by global models (see repository\footnote{\UrlFont{https://github.com/calgo-lab/resilient-timeseries-evaluation}}), such as Temporal Fusion Transformer (TFT) and Long Short-Term Memory (LSTM), which leverage aggregated information from the many sensors across the network~\cite{singh_data-driven_nodate}. Under network outage operation, the visualization switches to short-horizon edge-based forecasts computed solely from local basin filling-level using LSTM or Transformer architectures (see repository\footnote{\UrlFont{https://github.com/tianheng-ling/EdgeOverflowForecast
}}) optimized for deployment under constrained conditions~\cite{ling_automated_2025}. All models are trained to minimize the Mean Squared Error between the predictions and measurements. 

\paragraph{Rainfall Visualization and Forecasting.}

\begin{figure}[!tbp]
    \centering
    \includegraphics[width=\columnwidth]{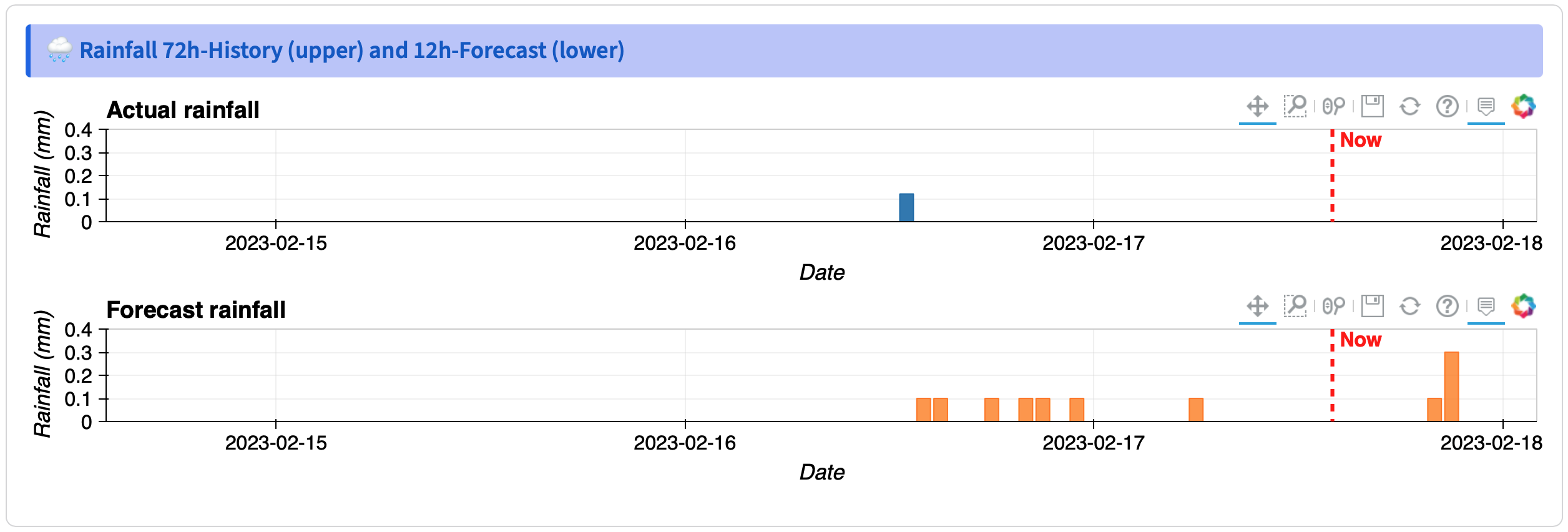}
    \caption{Rainfall observations and forecasts across conditions.}
    \label{fig:rainfall}
\end{figure}

To provide contextual information for filling-level forecasting, the dashboard includes a rainfall visualization module aligned with the forecast view (see \autoref{fig:rainfall}).  Historical and forecasted (when available) rainfall are shown as bar charts on a shared time axis, allowing users to relate precipitation dynamics to predicted basin filling behavior. Hover interaction enables inspection of exact precipitation amounts at each time step. Under standard operation, both historical and forecasted rainfall are displayed and incorporated into cloud-based long-horizon prediction. Under network outage operation, rainfall remains visualized, but is excluded from the forecasting process as external data is unavailable due to connectivity loss.

\begin{figure}[!tbp]
    \centering
    \includegraphics[width=.55\columnwidth]{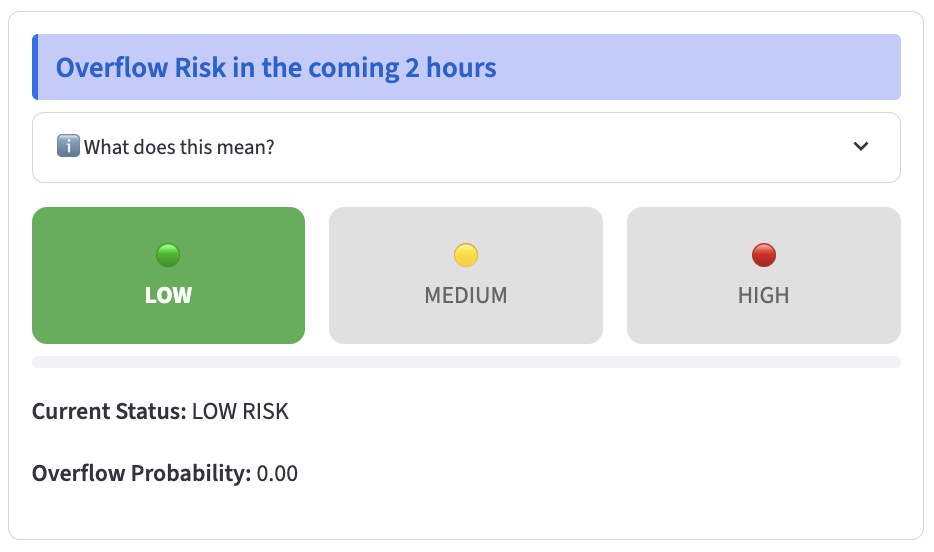}
    \caption{A traffic light indicator conveys overflow risk in next two hours derived from recent rainfall and current network conditions.}
    \label{fig:overflow_risk}
\end{figure}

\paragraph{Overflow Risk.}

The overflow risk module displays the estimated risk for the next two hours using a traffic-light indicator (see \autoref{fig:overflow_risk}). A gradient-boosted decision tree (see repository\footnote{\UrlFont{https://git.okeanos.ai/peter.ghaly/riwwer\_public}}) is used to output a confidence score mapping to three risk levels: low ($<$33\%, green), medium (33-67\%, yellow), and high ($>$67\%, red). This model was trained using the overflow occurrence after rainfall events, recorded by Wirtschaftsbetriebe Duisburg. Inspired by~\cite{jalbert2024categorization}, the estimation relies on engineered features derived from past 24 hours of rainfall measurements. The features include the daily maximum rainfall accumulation during different periods, complemented by the current filling-level. In addition, cyclonic features (such as the month, day, and hour) are also included to enhance the model performance.


\section{Conclusion and Future Work}

This study presents a resilient solution that consolidates cloud- and edge-based overflow forecasting into an interactive monitoring dashboard for urban sewer management. The system makes overflow risk and early-warning behavior observable under both conditions, highlighting operational trade-offs arising from changing connectivity and sensor availability. 
Future work will extend the demonstrator toward live data integration and broader robustness analysis, such as outliers or missing values.

\anon{}{
\section*{Acknowledgments}
The authors acknowledge financial support from the Federal Ministry of Research, Technology and Space of Germany for the RIWWER project (Grant Nos. 01MD22007H, 01MD22007C) and from the German Research Foundation (DFG) under Project No. 528483508 (FIP 12).
}

\bibliographystyle{named}
\bibliography{references}
\end{document}